%% file: iclr2027_conference_semtrace_general_method.tex
\documentclass{article} 
\usepackage{iclr2027_conference,times}

\input{math_commands.tex}

\usepackage{amsmath,amssymb}
\usepackage{hyperref}
\usepackage{url}
\usepackage{booktabs}

\title{SemTrace: Source-Grounded Semantic Signatures for Tracing LLM Exposure to Protected Documents}

\author{
Junyan Zhang$^{1}$,
Yudong Zeng$^{2}$,
Yongwei Huang$^{2}$,
Zuhao Ouyang$^{2}$,
Hong Chen$^{1}$,
Xuming Hu$^{1,*}$ \\ \\
$^{1}$The Hong Kong University of Science and Technology (Guangzhou), Guangzhou, China \\
$^{2}$Bosum Institute of Management Science, Shenzhen, China \\
$^{*}$Corresponding author
}

\iclrfinalcopy 
\begin{document}

\maketitle
\fancyhead{}

\begin{abstract}
Large language models are increasingly used to read documents and produce downstream text, creating a provenance problem when the document owner cannot control or inspect the model that performs the generation. We introduce \textsc{SemTrace}, a source-grounded semantic watermark for detecting whether a generated review was influenced by a known protected manuscript copy. Rather than biasing token probabilities or imposing surface-form patterns, \textsc{SemTrace} constructs a document-specific binary signature from factual propositions that are directly supported by the manuscript itself. A protected PDF invisibly carries a content contract that selects one fact from each binary pair and asks an instruction-following reviewer to express those facts in fixed review slots without changing its independent evaluation. A frozen natural language inference model then decodes the resulting semantic evidence with explicit erasures and scores the recovered bits against the codeword assigned to that copy. This design targets model-agnostic, assigned-copy exposure detection while keeping the watermark semantically tied to the source document.
\end{abstract}

\section{Introduction}
\label{sec:intro}

Large language models (LLMs) are rapidly becoming general-purpose readers and writers. A user can provide a long document---a paper, report, assignment, contract, or internal memo---and ask a model to summarize, critique, transform, or reason about it. This workflow is useful, but it creates a difficult provenance question for the document owner: \emph{if a downstream text was produced after an LLM read a particular protected copy of a document, can that exposure be detected from the generated text alone?} In many realistic settings, the document owner neither controls the LLM nor has access to its logits, decoding process, API provider, or model weights. Academic peer review is a representative example: a venue may distribute a manuscript to a reviewer, yet cannot assume access to whatever external model the reviewer may use to process it \citep{rao2025detecting, liu2026context}.

Most LLM watermarking methods address a different deployment setting. They embed statistical structure during model decoding, for example by biasing token selection toward a keyed subset of the vocabulary. Such methods have established that generated text can carry detectable signals with small quality loss, and later work has studied robustness, unbiased sampling, and adaptive watermark strength. However, these approaches generally require the party embedding the watermark to control generation \citep{kirchenbauer2023watermark, kuditipudi2023robust, liu2024adaptive}. This assumption breaks when a trusted party controls only the \emph{input document}, while generation is performed by an unknown third-party LLM.

In-context watermarking (ICW) provides a compelling alternative: instead of modifying the model, a trusted party places a watermarking instruction in the model's context and relies on instruction following to induce a detectable pattern in the output\citep{liu2026position}. Recent work demonstrates this idea with Unicode, word-initial, lexical, and acrostic signals, including an indirect-prompt-injection setting in which the instruction is embedded in a document \citep{greshake2023not, liu2026context}. This removes the need for decoding access, but the watermark signal itself is largely expressed through surface form. Such patterns are intentionally orthogonal to the document's substantive content: particular characters, initials, vocabulary items, or sentence initials are encouraged because they carry a key, not because they are facts supported by the source. This creates a natural tension between detectability, unobtrusiveness, instruction-following burden, and robustness to rewriting.

We explore a different design point: \emph{source-grounded semantic signatures}. Our method, \textsc{SemTrace}, constructs the watermark alphabet from the document itself. For each paper, we extract atomic propositions that are directly supported by auditable source passages, filter them for factuality, neutrality, and redundancy, and use unwatermarked reviews to estimate how often each fact would be mentioned naturally. We then select a set of reliable and relatively rare facts, arrange them into binary alternatives, and assign the document a balanced codeword that chooses one side of each pair. The protected copy contains an invisible instruction asking the review model to naturally express the selected facts in predetermined content slots. The facts are not fabricated claims: both alternatives in every pair are true statements supported by the paper. The signature therefore changes \emph{which true details are selected for expression}, rather than forcing arbitrary lexical artifacts into the review.

Detection is semantic rather than string based. Given a review, a frozen natural language inference (NLI) model evaluates whether either side of each fact pair is entailed by local review spans. Each position is decoded as 0, 1, or an explicit erasure when evidence is absent, conflicting, or ambiguous. The observed positions are compared only with the codeword assigned to the known protected copy. Accordingly, our primary task is \emph{assigned-copy exposure detection}: determining whether a review carries the semantic signature of a specified protected manuscript copy. The signature is a structured detection code rather than a claim of arbitrary-payload communication, and the single-copy main protocol does not require scanning a multi-recipient codebook.

The key technical challenge is that semantic carriers are not automatically reliable. A fact that appears frequently in ordinary reviews produces false evidence; two near-duplicate facts waste capacity; and if the selected fact semantically entails its unselected counterpart, the detector may observe both sides and erase the position. \textsc{SemTrace} therefore couples source verification, natural-mention estimation, diversity-aware selection, and code-aware pairing. Seed-disjoint unwatermarked reviews estimate each candidate's natural mention prior. A maximal-marginal-relevance objective favors source-supported, important, verifiable, naturally expressible, but control-rare facts. A subsequent joint assignment procedure chooses fact pairs, bit orientation, and review slots while explicitly penalizing selected-to-unselected semantic implication.

Our contributions are:
\begin{itemize}
    \item We introduce a source-grounded semantic signature that encodes a balanced binary codeword through alternative, manuscript-supported factual propositions rather than token-level or formatting patterns.
    \item We develop a fully auditable construction pipeline combining exact source quotations, source-NLI validation, neutrality constraints, semantic deduplication, control-aware MMR selection, and code-aware joint pairing and slot assignment.
    \item We introduce an erasure-aware NLI detector that recovers semantic bits from generated reviews and scores them against the known assigned codeword. Our evaluation protocol uses paired protected/unprotected generations, paper-level clustered uncertainty, matched ICW baselines, and black-box paraphrase stress tests.
\end{itemize}

\section{Related Work}
\label{sec:related}

\subsection{Watermarking LLM-Generated Text}

A large body of work embeds machine-detectable signals directly into LLM generations. A canonical approach modifies next-token sampling so that a secret key defines preferred ``green'' tokens and detection tests whether these tokens occur more often than expected by chance \citep{kirchenbauer2023watermark}. Follow-up work studies reliability under editing and mixing, distortion-free constructions, adaptive watermarking, security, and stronger statistical  \citep{kirchenbauer2024reliability, kuditipudi2023robust, liu2024adaptive, liu2025vla, huo2026pmark, dathathri2024scalable, liu2024semantic, zhang2025cohemark}. These methods are well suited to model providers that can intervene in decoding, but this access requirement is restrictive when watermarking must be initiated by a document owner rather than by the generator.

Post-hoc watermarking relaxes decoder access by transforming already generated text. For example, PostMark derives an input-dependent vocabulary through semantic representations and rewrites text to incorporate selected words \citep{chang2024postmark}. This setting is complementary to ours: a post-hoc watermarker controls the output after generation, whereas \textsc{SemTrace} assumes control only over the \emph{input document} before an unknown model produces the downstream text.

\subsection{Semantic Watermarks and Robustness to Paraphrasing}

Token-level watermarks can be weakened by sufficiently aggressive rewriting, motivating watermarks defined at coarser semantic granularity. SemStamp partitions sentence embedding space with locality-sensitive hashing and uses rejection sampling to place generated sentences in keyed semantic regions; k-SemStamp replaces random hyperplanes with clustering to improve the robustness--efficiency trade-off \citep{hou2024semstamp}. These methods define watermark states over representations of generated sentences. \textsc{SemTrace} uses ``semantic'' in a different sense: its symbols are explicit source-grounded propositions, and detection asks whether a downstream text entails one of two factual alternatives associated with each bit. This makes every decoded position interpretable in terms of a concrete claim from the protected document.

Robustness is also central to the broader literature on AI-text detection. Recursive paraphrasing and other rewriting attacks can substantially reduce the effectiveness of both learned detectors and watermark-based detectors, while reliability studies show that some watermark evidence can survive moderate human or machine rewriting when enough text is observed \citep{sadasivan2023can, kirchenbauer2024reliability, rastogi2024revisiting}. Motivated by this tension, our evaluation treats paraphrasing as an explicit stress test and distinguishes semantic survival from surface overlap.

\subsection{In-Context Watermarking and Document-Triggered Instructions}

In-context watermarking removes the need for privileged model access by expressing the watermark as an instruction in the model context. In the peer-review setting, \citet{rao2025detecting} embed hidden
instructions in manuscript PDFs to induce distinctive phrases,
technical terms, or citations in LLM-generated reviews and detect
their presence statistically. \citet{liu2026context} study Unicode, initials, lexical, and acrostic ICWs, and further instantiate indirect prompt injection by placing watermarking instructions into documents such as academic manuscripts. We share the same core deployment advantage---the protected-document owner can trigger a watermark without modifying the downstream LLM---but differ in the representation of the signal. Existing ICW variants primarily manipulate orthographic, lexical, or structural statistics. \textsc{SemTrace} instead binds each watermark position to a pair of true, source-supported facts and detects the resulting content choices with NLI.

The carrier mechanism is also related to indirect prompt injection, where instructions embedded in retrieved or externally supplied content are interpreted by an LLM as commands rather than passive data. Prior security work demonstrates that this data--instruction boundary can be exploited to influence downstream model behavior \citep{greshake2023not}. We use the same underlying capability for a narrowly scoped provenance mechanism: the protected copy contains a document-owner constraint whose intended effect is limited to selecting among source-supported content details, while explicitly instructing the reviewer not to alter its score, recommendation, criticism strength, or factual judgment.

\section{Method}
\label{sec:method}

\subsection{Problem Setting}
\label{sec:problem}

Let $d$ denote a source document and $y$ a downstream text generated after an LLM reads that document. The trusted party distributes a protected copy $d^{+}$ and retains the original unwatermarked copy $d^{-}$. The generation model is treated as a black box: the trusted party does not require access to model weights, logits, decoding, or provider-side watermarking infrastructure. Our goal is to test whether $y$ carries the semantic signature assigned to $d^{+}$.

For a configurable signature length $K$, \textsc{SemTrace} constructs $K$ binary fact pairs
\begin{equation}
\mathcal{P}_d = \left\{\left(f^{0}_{d,j}, f^{1}_{d,j}\right)\right\}_{j=1}^{K},
\end{equation}
where both $f^{0}_{d,j}$ and $f^{1}_{d,j}$ are independently true propositions supported by the document. The protected copy receives a document-specific binary codeword
\begin{equation}
\mathbf{c}_d = (c_{d,1},\ldots,c_{d,K}) \in \{0,1\}^{K}.
\end{equation}
We use balanced codewords to avoid a systematic preference for either side of the fact pairs. The codeword selects one fact from each pair,
\begin{equation}
 g_{d,j}=f^{c_{d,j}}_{d,j},
\end{equation}
and the carrier asks an instruction-following model to express the selected facts in predetermined content slots. At detection time, we recover
\begin{equation}
\hat{\mathbf c}(y)\in\{0,1,?\}^{K},
\end{equation}
where ``?'' denotes an erasure when the generated text does not provide sufficiently clear evidence for either side.

The main hypothesis test is deliberately narrow. Under $H_0$, the downstream text does not carry the signature of the known protected copy; under $H_1$, its recoverable semantic choices agree with the assigned codeword. We compare against the single assigned codeword rather than scanning a multi-recipient codebook. Thus, the binary vector serves as a structured exposure-detection signature rather than an arbitrary payload channel.

\subsection{Source-Grounded Atomic Fact Pool}
\label{sec:facts}

A semantic watermark is only meaningful if its carriers are faithful to the source. We therefore construct the fact pool from the document body while retaining explicit provenance for every candidate. The source is parsed into stable evidence units, and candidate propositions are generated only from evidence within the retained document scope. Each candidate is linked to a designated source unit and an exact supporting quotation.

Candidate facts pass a sequence of frozen gates. We require structural validity and exact quotation support, then verify that the proposition is entailed by a local source window using a frozen NLI model. We additionally exclude facts whose estimated sentiment or rating impact is too strong, because the watermark should select content details without deliberately steering the downstream evaluation. Finally, we remove redundant propositions using both representation-level similarity and bidirectional semantic entailment. If the approved pool is too small to construct all binary positions, the system searches uncovered evidence for additional candidates under the same gates rather than relaxing the acceptance criteria. If a valid pool still cannot be formed, construction fails explicitly for that document.

Formally, let $e_f$ denote source entailment, and let $a_f$ collect auxiliary attributes such as importance, verifiability, natural expressibility, sentiment, and rating impact. Candidate acceptance is written abstractly as
\begin{equation}
 f\in\mathcal{F}_d
 \quad\Longleftrightarrow\quad
 e_f\geq \tau_{\mathrm{src}}
 \ \land\ 
 \mathrm{Neutral}(a_f)
 \ \land\ 
 \mathrm{Distinct}(f;\mathcal{F}_d),
\end{equation}
where the thresholds and concrete duplicate tests are fixed before evaluation and reported with the experimental configuration.

\subsection{Estimating Natural Mention Priors}
\label{sec:controls}

Not every true fact is a useful watermark carrier. Central contributions may be mentioned by ordinary downstream texts even without the protected carrier, making their presence weak evidence of exposure. To estimate this background rate, we generate a separate set of \emph{selection-only controls} from the original unwatermarked document. These controls use a seed space disjoint from the final evaluation and are used only during signature construction.

Let $N_{\mathrm{sel}}$ be the number of selection controls and $q_{f,k}$ the NLI coverage of candidate fact $f$ in control $k$. With mention threshold $\tau_{\mathrm{mention}}$ and add-one smoothing, we estimate
\begin{equation}
 r_f =
 \frac{\sum_{k=1}^{N_{\mathrm{sel}}}\mathbb{1}[q_{f,k}\geq \tau_{\mathrm{mention}}]+1}
 {N_{\mathrm{sel}}+2}.
\end{equation}
A smaller $r_f$ indicates that the fact is less likely to appear spontaneously and is therefore potentially more discriminative. Selection-only controls are never reused as final negative examples.

\subsection{Selecting Reliable and Nonredundant Facts}
\label{sec:mmr}

To instantiate $K$ binary positions, we select $2K$ facts from the approved pool. Each fact receives a utility score that combines source support, importance, verifiability, natural expressibility, and rarity under unwatermarked controls:
\begin{equation}
B(f)=
\alpha_e e_f+\alpha_i i_f+\alpha_v v_f+\alpha_n n_f+\alpha_r(1-r_f),
\end{equation}
where the $\alpha$ coefficients are fixed hyperparameters.

We also penalize redundancy between candidates. Let $\mathbf q_f$ denote the vector of control-coverage scores for $f$, and let $J_{\mathrm{lex}}(f,g)$ denote lexical overlap. We use a generic redundancy function
\begin{equation}
R(f,g)=
\beta_c\max\!\left(0,\mathrm{corr}(\mathbf q_f,\mathbf q_g)\right)
+\beta_l J_{\mathrm{lex}}(f,g).
\end{equation}
Starting from an empty set $S$, greedy maximal marginal relevance selects the next fact maximizing
\begin{equation}
B(f)-\lambda_{\mathrm{MMR}}\max_{g\in S}R(f,g),
\end{equation}
until $2K$ facts are chosen \citep{carbonell1998use}. This favors individually strong carriers while discouraging semantically or behaviorally redundant positions.

\subsection{Code-Aware Pairing, Orientation, and Slot Assignment}
\label{sec:assignment}

The selected $2K$ facts must be converted into $K$ binary positions. A naive construction can be fragile: if the code-selected side of a pair semantically entails its unselected counterpart, the detector may find evidence for both facts and erase the position. We therefore jointly optimize fact pairing, 0/1 orientation, and content-slot assignment with knowledge of the document-specific codeword.

For candidate facts $f$ and $g$, the pairing cost combines mismatch in their natural mention priors and content attributes, their tendency to co-occur in selection controls, semantic redundancy, and optional category mismatch:
\begin{align}
C_{\mathrm{pair}}(f,g) ={}&
\lambda_r |r_f-r_g| + \lambda_i |i_f-i_g| + \lambda_n |n_f-n_g| \\
&+\lambda_{\mathrm{co}} C_{\mathrm{co}}(f,g)
+\lambda_R R(f,g)
+\lambda_{\mathrm{cat}}\mathbb{1}[\mathrm{cat}(f)\neq\mathrm{cat}(g)].
\end{align}
Here $C_{\mathrm{co}}(f,g)$ measures how often the two candidates are simultaneously covered in selection controls.

Let $E(f\rightarrow g)$ denote directed semantic entailment between two candidate facts. We transform entailment above a reference threshold into an implication-risk function $\rho(f\rightarrow g)$. The complete assignment objective takes the form
\begin{equation}
\mathcal{L}=\lambda_A\mathcal{A}+\lambda_P\mathcal{P}+\lambda_H\mathcal{H},
\end{equation}
where $\mathcal{A}$ aggregates implication risk from code-selected facts to their unselected counterparts, $\mathcal{P}$ is the total pair cost, and $\mathcal{H}$ captures preferences induced by the downstream content slots. We initialize with a control-aware matching and refine the joint assignment using a deterministic search procedure. The optimization therefore determines which facts form a bit, which side is 0 or 1, and where the selected fact should appear in the generated text.

\subsection{Document Carrier}
\label{sec:carrier}

After assignment, the codeword selects one fact $F_j$ for every semantic position. The carrier maps these selected facts to predetermined content slots rather than leaving the model free to distribute them arbitrarily. The instruction permits faithful paraphrasing but requires each selected fact to remain independently recognizable and to preserve meaning-critical details such as named entities, datasets, comparisons, numbers, and causal qualifiers.

Crucially, selected facts serve only as evidence or context. The carrier instructs the downstream model not to change its independent score, recommendation, criticism strength, or factual judgment; not to invent unsupported results; and not to reveal the watermark, fact identifiers, or slot mapping. The exact slot layout is part of the experimental instantiation rather than the general definition of \textsc{SemTrace}.

The carrier is placed in an invisible text layer of the protected document so that visible content remains unchanged. After construction, the embedded instruction is re-extracted and checked against the intended text under a narrowly defined normalization. We also verify carrier boundaries and integrity metadata. A failed round trip causes construction to abort rather than silently using a corrupted carrier.

\subsection{Semantic Bit Decoding with NLI}
\label{sec:decoder}

The detector receives only the generated text $y$ and the precomputed fact pairs for the known document; it does not read the carrier instruction itself. We split $y$ into local windows $\mathcal{W}(y)$ and, for each position $j$ and side $b\in\{0,1\}$, compute
\begin{equation}
s_{j,b}(y)=
\max_{w\in\mathcal{W}(y)}
P_{\mathrm{NLI}}\!\left(f^{b}_{d,j}\mid w\right).
\end{equation}

We decode conservatively with explicit erasures. Let $\tau_{\mathrm{both}}$, $\tau_{\mathrm{min}}$, and $\tau_{\mathrm{margin}}$ be frozen thresholds. Then
\begin{equation}
\hat c_j =
\begin{cases}
?, & s_{j,0}\geq\tau_{\mathrm{both}} \ \land\ s_{j,1}\geq\tau_{\mathrm{both}},\\
?, & \max(s_{j,0},s_{j,1})<\tau_{\mathrm{min}},\\
?, & |s_{j,1}-s_{j,0}|\leq\tau_{\mathrm{margin}},\\
\mathbb{1}[s_{j,1}>s_{j,0}], & \text{otherwise.}
\end{cases}
\end{equation}
The three erasure cases respectively handle evidence for both alternatives, absent evidence, and insufficient semantic separation.

Let $O$ be the number of non-erased positions, $M$ the number matching the assigned codeword, and $U=O-M$ the number of mismatches. For an eligibility threshold $O_{\min}$, the assigned-code score is
\begin{equation}
S(y)=
\begin{cases}
M-U=2M-O, & O\geq O_{\min},\\
s_{\mathrm{inel}}, & O<O_{\min},
\end{cases}
\end{equation}
where $s_{\mathrm{inel}}$ is fixed below the minimum score attainable by an eligible codeword. Erased positions contribute zero to the signed match count. This prevents low-coverage texts from appearing positive merely because a small number of observed positions happen to agree.

\section{Experiments}
\label{sec:experiments}

We evaluate \textsc{SemTrace} in the academic peer-review setting using papers from ICML 2023 and ICLR 2024. We instantiate the general framework with a frozen Exact-12 configuration and keep the watermark construction, review generation, semantic decoding, and evaluation protocol fixed across the two corpora. We first describe the paper corpora, the Exact-12 instantiation, and the paired detection protocol, and then report results on clean detection and robustness to black-box paraphrasing.

\subsection{Paper Corpora}
\label{sec:corpora}

We construct our evaluation corpus from accepted papers at ICML 2023 and ICLR 2024. These two conferences provide diverse and technically detailed research papers spanning a broad range of machine learning topics, making them suitable for evaluating document-level semantic watermarking under realistic academic-review settings. We use the original paper PDFs as the source documents for all subsequent fact extraction, watermark construction, review generation, and detection experiments.

\subsection{Exact-12 Configuration}
\label{sec: configuration}

We use $K=12$ binary positions, selecting $2K=24$ approved facts per paper and assigning a balanced codeword with six zeros and six ones. Eight seed-disjoint selection-only controls are used to estimate the natural mention tendency of candidate facts.

Candidate facts are generated with Qwen2.5-VL-32B-Instruct and filtered for source support, evaluative neutrality, and semantic redundancy. We then select 24 complementary facts and pair them into 12 binary positions using the utility, redundancy, and control-aware objectives defined in Sections~\ref{sec:mmr} and~\ref{sec:assignment}. All thresholds, candidate-generation settings, selection weights, and pairing optimization details are fixed across experiments and are provided in the Appendix.

\subsection{Carrier and Review Generation}
\label{sec:reviewgen}

The 12 selected facts are mapped to a fixed conference-review layout: four facts occupy the four Summary sentences; two facts are assigned to the first Strength, one to each of the second and third Strengths; two to the first Weakness, and one to each of the second and third Weaknesses. We deliberately adopt this structured review format so that each semantic position has a predictable content location. The slot contract reduces fact omission and clustering, limits cross-position interference during NLI decoding, and makes the resulting semantic evidence easier to audit and compare across paired generations. It is therefore an experimental control for reliable semantic transmission and measurement rather than a general requirement of \textsc{SemTrace}.

The complete watermark instruction is embedded invisibly in the protected PDF, and reviews are generated using Qwen2.5-VL-32B-Instruct. Protected and unprotected generations use identical model and review-format constraints, so that the paired comparison differs in the presence of the document-triggered carrier rather than in the outer review format. Full carrier-construction and generation parameters are provided in Appendix~\ref{app:generation}.

\subsection{Detection and Evaluation Protocol}
\label{sec: detection}

For detection, we use a frozen DeBERTa-v3-small NLI model to determine which member of each fact pair is supported by the generated review. Reviews with at least 8 of the 12 positions observed are eligible for assigned-code scoring. The detector and all decision rules are fixed across datasets and experimental conditions; detailed NLI thresholds and decoding parameters are provided in the appendix.

For each paper, we generate five paired positive/null review draws. The positive member receives the protected PDF containing the carrier, whereas the null member receives the original unwatermarked PDF. Positive and null reviews within each pair share the same sampling seed. Selection-only controls use a separate seed space and are used only during watermark construction, not as negative examples in the final evaluation.

Our primary detection metric is ROC-AUC over the assigned-code score. We additionally report empirical zero-FPR and 5\% FPR operating points as descriptive summaries. Because multiple review draws from the same paper are not independent, confidence intervals are estimated using cluster bootstrap over papers while retaining all within-paper draws. The same paired evaluation protocol is used for the lexical, initials, and acrostic ICW baselines \citep{liu2026context}.

We further evaluate robustness under two black-box paraphrase attacks: a neutral paraphrase that rewrites wording and structure while preserving the original meaning and evaluation, and an adaptive scheme-agnostic paraphrase in which the attacker is informed that an unknown watermark may be present but receives no information about the paper-specific fact pairs, codeword, detector scores, or key. The detector remains frozen under both attacks.

\subsection{Experiment Results}
\label{sec:preliminary-results}

We evaluate \textsc{SemTrace} on the frozen ICML 2023 and ICLR 2024 corpora under the paired protocol described above. These preliminary results use the same construction, generation, detection, and evaluation pipeline as the main experiment.

\paragraph{Clean detection.}
\textsc{SemTrace} yields near-perfect separation between protected and unprotected reviews on both corpora. On ICML 2023, the assigned-code score achieves a ROC-AUC of $1.000$ with a paper-cluster bootstrap 95\% CI of $[1.000,1.000]$. On ICLR 2024, the corresponding ROC-AUC is $0.990$ with a 95\% CI of $[0.970,1.000]$. At the descriptive empirical zero-FPR operating point, TPR is $1.00$ on ICML and $0.98$ on ICLR.

Protected reviews expose substantially more decodable semantic positions than matched controls. The mean numbers of observed bits are $10.76$ versus $1.46$ on ICML and $10.84$ versus $1.26$ on ICLR. Among non-erased positive bits, agreement with the assigned codeword is $1.0000$ and $0.9945$, respectively. The mean paired positive-minus-null score differences are $+23.76$ for ICML and $+23.34$ for ICLR.

\begin{table}[t]
\centering
\small
\begin{tabular}{lcc}
\toprule
 & ICML 2023 & ICLR 2024 \\
\midrule
ROC-AUC & 1.000 & 0.990 \\
95\% cluster CI & [1.000, 1.000] & [0.970, 1.000] \\
Zero-FPR TPR & 1.00 & 0.98 \\
Positive observed bits & 10.76 & 10.84 \\
Control observed bits & 1.46 & 1.26 \\
Positive bit accuracy & 1.0000 & 0.9945 \\
Paired score difference & +23.76 & +23.34 \\
\bottomrule
\end{tabular}
\caption{Clean detection results for \textsc{SemTrace}. The zero-FPR operating point is measured descriptively on the evaluation data rather than using an independently calibrated threshold.}
\label{tab:preliminary-clean}
\end{table}

\paragraph{Paraphrase robustness.}
The semantic signal remains detectable after both black-box paraphrase attacks, although robustness depends on the rewriting strategy. Under neutral paraphrasing, ROC-AUC decreases to $0.880$ on ICML and $0.960$ on ICLR. At the frozen clean operating threshold, the corresponding TPRs are $0.68$ and $0.92$, with no observed false positives.

The adaptive unknown-scheme attack is less damaging in the current evaluation, yielding ROC-AUCs of $0.990$ on ICML and $0.980$ on ICLR and clean-threshold TPRs of $0.98$ and $0.94$, respectively. Protected reviews retain on average $9.06$ and $9.50$ observed bits after neutral paraphrasing, compared with $10.18$ and $10.64$ under the adaptive attack. Non-erased bit accuracy remains high in all four conditions, ranging from $0.9845$ to $1.0000$.

The apparently weaker adaptive attack should not be interpreted as robustness to a knowledgeable white-box adversary. The attacker is not given the source paper, fact pairs, assigned codeword, detector scores, or key. These results therefore characterize strong robustness to the evaluated scheme-agnostic black-box rewriting conditions.

\begin{table}[t]
\centering
\small
\begin{tabular}{llccc}
\toprule
Dataset & Attack & ROC-AUC & TPR@clean thr. & Obs. bits \\
\midrule
ICML 2023 & Neutral  & 0.880 & 0.68 & 9.06 \\
ICML 2023 & Adaptive & 0.990 & 0.98 & 10.18 \\
ICLR 2024 & Neutral  & 0.960 & 0.92 & 9.50 \\
ICLR 2024 & Adaptive & 0.980 & 0.94 & 10.64 \\
\bottomrule
\end{tabular}
\caption{Robustness results under neutral and adaptive unknown-scheme paraphrasing. TPR is evaluated using the frozen clean operating threshold.}
\label{tab:preliminary-paraphrase}
\end{table}




\bibliography{iclr2027_conference}
\bibliographystyle{iclr2027_conference}

\appendix

\section{Implementation Details}
\label{app:implementation}

This appendix provides the frozen implementation details omitted from the main experimental setup for readability. All configurations described below are fixed across datasets and experimental conditions.

\subsection{Exact-12 Construction Details}
\label{app:exact12}

For the Exact-12 configuration, we use $K=12$ binary positions and select $2K=24$ approved facts per paper. The assigned codeword is balanced, containing six zeros and six ones. Eight seed-disjoint selection-only controls are generated to estimate the natural mention tendency of candidate facts. Natural mention rates are computed using add-one smoothing, and the mention threshold is set to $0.52$.

Candidate generation uses Qwen2.5-VL-32B-Instruct over approximately 15,000-character evidence chunks with a three-unit overlap. Each chunk receives an initial candidate-generation pass followed by a supplemental pass. Candidate facts are constrained to 12--35 words and must include an exact supporting quotation of at least five words from the source document.

Source support is verified using NLI, with a minimum entailment probability of $0.68$. To reduce evaluative bias, we retain only candidates whose absolute sentiment and rating-impact scores are at most $0.35$. Near-duplicate candidates are removed at an embedding cosine similarity of $0.99$. Semantic-equivalence deduplication additionally uses bidirectional entailment with threshold $0.75$ together with lexical Jaccard similarity threshold $0.35$. If fewer than 24 candidates survive these filters, targeted refill searches are performed over uncovered evidence without relaxing any filtering threshold.

For fact selection, the base utility of candidate $f$ is
\begin{equation}
B(f)=1.2e_f+0.8i_f+0.5v_f+0.5n_f+1.0(1-r_f),
\end{equation}
and pairwise redundancy is defined as
\begin{equation}
R(f,g)=0.6\max\!\left(0,\mathrm{corr}(\mathbf q_f,\mathbf q_g)\right)
+0.4J_{\mathrm{lex}}(f,g).
\end{equation}
Greedy MMR selection uses coefficient $1.2$.

The frozen pairing cost is
\begin{align}
C_{\mathrm{pair}}(f,g) ={}&
3\left(|r_f-r_g|+|i_f-i_g|+|n_f-n_g|\right) \\
&+2C_{\mathrm{co}}(f,g)+2R(f,g)
+0.2\mathbb{1}[\mathrm{cat}(f)\neq\mathrm{cat}(g)].
\end{align}
Directed implication risk is activated at entailment $0.52$, and the assignment objective is
\begin{equation}
8\mathcal{A}+\mathcal{P}+\mathcal{H}.
\end{equation}
We initialize pairing with control-aware minimum-weight perfect matching, refine the assignment using deterministic simulated annealing, and finish with greedy improving swaps.

\subsection{Carrier and Review Generation Details}
\label{app:generation}

The complete document-triggered instruction is placed on the final page of the protected PDF using a Unicode-capable font in invisible render mode. After construction, we verify the embedded instruction through a text round-trip check.

Review generation uses Qwen2.5-VL-32B-Instruct in bfloat16 with temperature $0.35$, top-$p=0.9$, and a maximum of 1,800 output tokens. Generated reviews are constrained to 400--1,200 words and contain the sections Summary, Strengths, Weaknesses, Questions for the authors, and Overall score. Protected and unprotected generations use the same review structure, formatting constraints, model, and generation parameters.

\subsection{Detection Details}
\label{app:detection}

Detection uses a frozen DeBERTa-v3-small NLI model. Each generated review is divided into adjacent two-sentence windows, with at most 128 windows retained for inference. The minimum entailment threshold is $0.52$, the ambiguity margin is $0.20$, and the threshold for identifying both members of a fact pair as mentioned is $0.90$.

A binary position is treated as observed only when the NLI-based decoding rule resolves sufficient evidence for one member of the corresponding fact pair. Reviews with at least 8 of the 12 positions observed are eligible for assigned-code scoring. Reviews that do not satisfy this minimum-observability requirement receive the fixed ineligible score $-13$, which places them below the eligible score range.

For every paper, we generate five paired positive/null review draws. Within each draw, the positive review is generated from the protected PDF and the null review from the original unwatermarked PDF. The two generations share the same sampling seed. Selection-only controls occupy a disjoint seed space and are used only during watermark construction; they are never included as negative examples in the final detector evaluation.

The primary detection metric is ROC-AUC over the assigned-code score. We additionally report empirical zero-false-positive and 5\% false-positive operating points as descriptive summaries rather than independently calibrated deployment thresholds. Because multiple review draws from the same source paper are statistically dependent, confidence intervals are estimated by cluster bootstrap over papers while retaining all review draws belonging to each sampled paper.

Under paraphrase evaluation, the detector and the clean operating threshold remain frozen. The neutral attack rewrites wording and structure while preserving the review's meaning and evaluation. In the adaptive scheme-agnostic condition, the attacker is informed only that the review may contain an unknown watermark; it is not given the source paper, selected fact pairs, assigned codeword, detector scores, or secret key.


\end{document}

%% file: math_commands.tex
\usepackage{amsmath,amsfonts,bm}

\def\eqref#1{equation~\ref{#1}}

\def\1{\bm{1}}

\DeclareMathAlphabet{\mathsfit}{\encodingdefault}{\sfdefault}{m}{sl}
\SetMathAlphabet{\mathsfit}{bold}{\encodingdefault}{\sfdefault}{bx}{n}

